\documentclass[11pt]{article}

\usepackage[]{emnlp2021}

\usepackage{times}
\usepackage{latexsym}

\usepackage[T1]{fontenc}

\usepackage[utf8]{inputenc}

\usepackage{microtype}

\usepackage{xcolor}
\usepackage{booktabs}
\usepackage{todonotes}
\usepackage{amssymb}
\usepackage{graphicx}
\usepackage{multirow}
\usepackage{mwe}
\usepackage{bbm}
\usepackage[normalem]{ulem}

\graphicspath{ {./images/} }

\newcommand{\mbert}[0]{M-BERT}
\newcommand{\bertmono}[0]{BERT+AKK(mono)}
\newcommand{\repourl}[0]{www.github.com/SLAB-NLP/Akk}
\newcommand{\example}[1]{\emph{``#1''}}
\newcommand{\hit}[1]{Hit@#1}

\title{Filling the Gaps in Ancient Akkadian Texts:\\ A Masked Language Modelling Approach}

\author{
 Koren Lazar$^{\clubsuit\diamondsuit}$\thanks{$\;\;$Work performed while at The Hebrew University of Jerusalem.} $\;\;$
 Benny Saret$^{\spadesuit}$ $\;\;$
 Asaf Yehudai$^{\diamondsuit}$  \\
 \bf
 Wayne Horowitz$^{\spadesuit}$ $\;\;$
 Nathan Wasserman$^{\spadesuit}$ $\;\;$
 Gabriel Stanovsky$^{\diamondsuit}$ \\
 $^{\clubsuit}$IBM Research \\ $^{\diamondsuit}$School of Computer Science and Engineering, The Hebrew University of Jerusalem \\$\;\;$ $^{\spadesuit}$The Institute of Archaeology, The Hebrew University of Jerusalem \\
 koren.lazar@ibm.com, gabriel.stanovsky@mail.huji.ac.il
}

\date{}
\begin{document}
\maketitle

\begin{abstract}
We present models which complete missing text given transliterations of ancient Mesopotamian documents, originally written on cuneiform clay tablets (2500 BCE - 100 CE).
Due to the tablets' deterioration, scholars often rely on contextual cues to manually fill in missing parts in the text in a subjective and time-consuming process.
We identify that this challenge can be formulated as a masked language modelling task, used mostly as a pretraining objective for contextualized language models. Following, we develop several architectures focusing on the Akkadian language, the lingua franca of the time. We find that despite data scarcity (1M tokens) we can achieve state of the art performance on missing tokens prediction (89\% hit@5) using a greedy decoding scheme and pretraining on data from other languages and different time periods.
Finally, we conduct human evaluations showing the applicability of our models in assisting experts to transcribe texts in extinct languages.
\end{abstract}

\section{Introduction}
\begin{figure}
    \centering
    \includegraphics[width=0.8\linewidth]{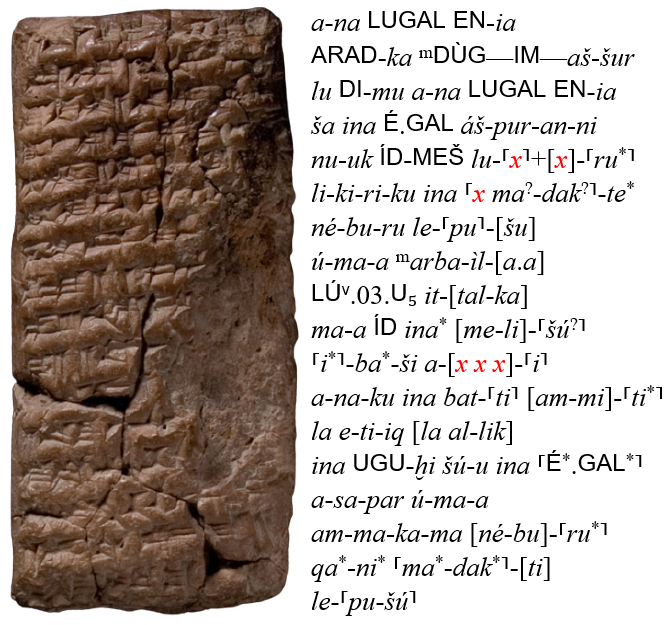}
    \caption{\label{fig:tablet+transliteration}A clay tablet from Oracc (left) with its corresponding Latin transliteration (right). Words are delimited by spaces, while signs are delimited by hyphens or dots. A sign which is missing due to deterioration is denoted by `x' and highlighted in red in the figure. We develop models which automatically complete these missing signs based on the surrounding context.
    }
\end{figure}

The Akkadian language was the lingua franca of the Middle East and Egypt in the Late Bronze and Early Iron Ages, spoken or in use from 2500 BCE until its gradual extinction around 100 CE~\citep{oppenheim2013ancient}. It was written in cuneiform signs --- wedge-shaped imprints on clay tablets, as depicted in Figure~\ref{fig:tablet+transliteration}~\citep{walker1987cuneiform}.
These tablets are the main record from the  Mesopotamian cultures, including religious texts, bureaucratic records, royal decrees, and more. Therefore they are a target of extensive transcription and transliteration efforts. One such transcription is exemplified by the Latinized text to the right of the tablet in Figure~\ref{fig:tablet+transliteration}. 

The Open Richly Annotated Cuneiform Corpus (Oracc)\footnote{\url{http://oracc.org}} is one of the major Akkadian transcription collections, culminating in approximately $2.3M$ transcribed signs from $10K$ tablets.
As further evidenced in Figure~\ref{fig:tablet+transliteration}, many of the signs in the tablets were eroded over time and some parts were broken or lost, forcing editors to ``fill in the gaps'' where possible, based on the context of the surrounding words.

In this paper, we identify that the task of masked language modeling, used ubiquitously in recent years for pretraining other downstream tasks~\citep{peters-etal-2018-deep,howard-ruder-2018-universal,liu2019roberta} lends itself directly to missing sign prediction in the transliterated texts. We experiment with various adaptations of BERT-based models~\citep{devlin2019bert} trained and tested on Oracc, combined with a greedy decoding scheme to extend the prediction from single tokens to multiple words. We specifically focus on the effect multilingual pretraining has on downstream performance, which was recently shown beneficial for low-resource settings~\citep{chau2020parsing}.

In an automatic evaluation, we find that a combination of large-scale multilingual pretraining with Akkadian finetuning achieves state-of-the-art results, with a top 5 accuracy of 89.5\%, vastly improving over other models and baselines.
Interestingly, we find that the multilingual pretraining signal seems to be more important than the signal of the target small-scale Akkadian data, as
the \emph{zero-shot} performance of a multilingual language model surpasses that of a monolingual Akkadian model by about 10\%.

Finally, we show the model's potential applicability in assisting transcription by filling in missing parts. To account for the challenges in human assessment of an extinct language, we created a controlled setup where domain experts are asked to identify plausible predictions out of a combination of model predictions, the original masked sequences, and noise. We find that in a majority of cases, the annotators found at least one of the model's top 3 predictions useful, while the performance degrades on longer sequences.
Future work can improve the model by designing more elaborate decoding schemes and exploring the specific effect of related languages (e.g., Arabic and Hebrew) on downstream performance.
Our code and trained models are made publicly available at \url{\repourl}.
    
Our main contributions are:
    \begin{itemize}
    \item We identify that the longstanding challenge of filling in gaps in Akkadian texts directly corresponds to advances in masked language modeling.
    \item We train the first Akkadian language model, which can serve as a pretrained starting point for other downstream tasks such as Akkadian morphological analysis.
    \item We develop state-of-the-art models for completing missing signs by combining large-scale multilingual pretraining with Akkadian language finetuning.
    \item We devise a controlled user study, showing the potential applicability of our model in assisting scholars fill in gaps in real-world Akkadian texts.
    
     \end{itemize}

\section{Background}
In this section, we will introduce the Akkadian language and the Open Richly Annotated Cuneiform Corpus (Oracc). While it is one of the largest sources of the Akkadian language, it is of orders of magnitude smaller compared to resources for other languages, such as English or German. Then, we will introduce masked language modeling, which will serve as the basis for our sign prediction model. 

\subsection{The Akkadian Language and the Oracc Dataset}
\label{subsec:Akkadian+Oracc}
Akkadian is a Semitic language, related to several languages spoken today, such as Hebrew, Aramaic, Amharic, Maltese, and Arabic. It has been documented from the 3\textsuperscript{rd} millennium B.C.E. until the first century of the common era, in modern Iraq, between the Euphrates and the Tigris rivers, as well as in modern Syria, east Turkey, and the Northern Levant~\cite{Huehnergard2011Akkadian}.
In this work, we will use the Open Richly Annotated Cuneiform Corpus (Oracc), one of the largest international cooperative projects gathering cuneiform texts from many archaeological sites.

\begin{table}[b!]
\centering
\begin{tabular}{llll}
\toprule
               & \# Texts & \# Words & \# Signs \\ \midrule
Akkadian Train & 8K       & 950K     & 1.8M     \\
Akkadian Test    & 2K       & 250K     & 500K     \\\midrule
English Train  & 7K       & 950K     & --       \\
English Test   & 2K       & 250K     & --      \\\bottomrule
\end{tabular}
\caption{Number of texts, words, and signs in our preprocessed version of Oracc, English texts are corresponding translations of the Akkadian texts. 
}
\label{tab:amounts-table}
\end{table}

Most relevant to this work, Oracc contains Latinized transliterations of the cuneiform texts, as can be seen in Figure~\ref{fig:tablet+transliteration}, depicting a clay tablet and its transliteration in Oracc. It also contains English translations for parts of the texts. In total, as can be seen in Table~\ref{tab:amounts-table}, Oracc consists of about $10K$ texts (each a transliteration of a single tablet), containing $1M$ words and $2.3M$ signs, as well as $9K$ translated texts in English containing $1.2M$ English words. 
Importantly, the editors can often visually estimate the number of missing signs in a deteriorated or missing part and denote each with `x' in the transliteration (marked in red in Figure~\ref{fig:tablet+transliteration}).
Therefore, in the following sections, we will assume that the number of missing signs is given as input to our models.


\subsection{Multilingual Masked Language Modeling}
\label{subsec:MLM}
In masked language modeling (MLM), a model is asked to predict masked parts in a text given their surrounding context.
Recent years have seen large gains for almost all NLP tasks by using the token representations learned during MLM as a starting point for downstream applications.
In particular, recent work has noticed that joint training on various languages greatly helps downstream applications, especially where labeled data is sparse~\citep{pires-etal-2019-multilingual,chau2020parsing,conneau-etal-2020-emerging}.

In this work we identify that the MLM objective directly corresponds to the task of filling in gaps in Akkadian texts and train several MLM variants on it. In the following sections, we will especially examine the effect of multilingual pretraining on our task.

\section{Task Definition}
\begin{figure}[b!]
    \centering
    \includegraphics[width=1.0\linewidth]{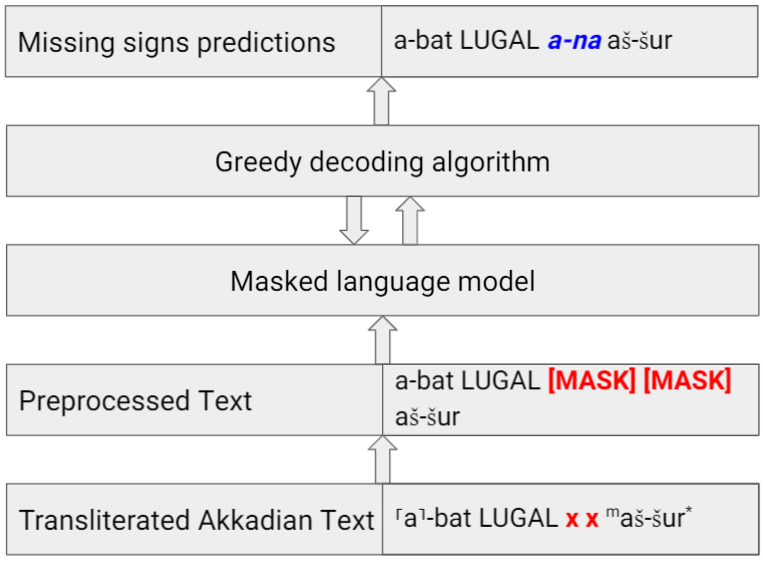}
    \caption{High-level diagram of our model, producing a sequence of signs (marked in blue) given input from Oracc with missing signs (red `x's). We experiment with different language models and pretraining data.
    }
    \label{fig:pipeline_figure}
\end{figure}

Intuitively, our task, as demonstrated in Figure~\ref{fig:pipeline_figure}, is to predict missing tokens or signs given their context in transliterated Akkadian documents. Human experts achieve this when compiling Oracc by considering not only the surrounding context in the tablet, but also its wider, external context, such as its corpus, or the time and location where the text was originally written or found. 
In many cases, researchers can estimate the number of missing signs even after their physical deterioration, and mark them as sequences of `x's. E.g., note the sequence of 2 `x's marked in red in  Figure~\ref{fig:pipeline_figure}. We will use this signal as input to our model, which specifies the number of signs to be predicted.\footnote{We filter cases where the editors can not estimate the number of missing signs.}

Formally, let $T=(s_1,...,s_n)\in\Sigma^n$ be a transliterated Akkadian document comprised of a concatenation of $n$ signs, where $\Sigma$ is the set of all Akkadian signs. Let $I\subseteq[n]$ such that $\forall i\in I:s_i=x$, where $x$ denotes a missing sign. The number of missing signs is assumed to be known a priori, based on the editor's examination of the tablets. Therefore, the model should output $(p_1,...,p_{|I|})\in\Sigma^{|I|}$ predictions for the missing signs in $T$.

\section{Model}
In this section, we will introduce BERT-based models aiming to solve the task of predicting missing signs in Akkadian texts.
We chose these models since their pretraining task is also our downstream task.
The high-level diagram of the model is presented in Figure~\ref{fig:pipeline_figure} and is elaborated below.
First, in Section~\ref{subsec:preprocessing}, we outline the preprocessing of Oracc, aiming to remove annotations that are external to the original text.
Then in Section~\ref{subsec:Masked language models}, we propose two models for predicting missing signs. Lastly, in Section~\ref{subsec:From Tokens to Signs}, we present an algorithm to extend BERT sub-word level prediction to multiple signs and words.
In the following two sections we will test these models in both automatic and human evaluation setups.

\subsection{Preprocessing}
\label{subsec:preprocessing}
Oracc is a collaborative effort to transliterate Mesopotamian tablets, mainly in Akkadian. Figure~\ref{fig:tablet+transliteration} exemplifies different characteristics of the corpus. 
We removed signs added by editors in the transliteration process as they were not part of the original text. For example, we removed signs which indicate how certain the editors are in their reading of the tablet.
As an example, note that in Figure~\ref{fig:pipeline_figure} the first sign in the transliterated text is marked as uncertain with the $\ulcorner\urcorner$ characters before preprocessing.
In addition, we also remove superscripts and subscripts, which indicate different readings of the Akkadian cuneiform text, e.g., an `$^m$' superscript is preceding the last word in the transliterated text.

During training, similarly to \citet{devlin2019bert}, we train the model to predict known tokens by masking them at random. 
During inference, we mask each missing sign, indicated by `x' in Oracc, and iteratively predict each of the tokens composing it.

\subsection{Masked Language Models}
\label{subsec:Masked language models}
We experimented with monolingual and multilingual versions of BERT.

First, we pretrained from scratch a monolingual BERT model with a reduced number of parameters ($750K$) following conclusions from \citet{kaplan2020scaling}.
Second, following recent research suggesting that pretraining on similar languages is beneficial for many NLP tasks, including in low-resource settings~\citep{pires-etal-2019-multilingual,wu-dredze-2019-beto,chau2020parsing,conneau-etal-2020-emerging}, we finetuned a pretrained multilingual BERT (\mbert{}) model \cite{devlin2019bert}.\footnote{https://huggingface.co/bert-base-multilingual-cased}
\mbert{} was trained on the 104 most common languages of Wikipedia, including Hebrew and Arabic - Semitic languages that are typologically similar to Akkadian.

To adapt \mbert{} to Akkadian, we assign its 99 available free tokens, optimizing for maximum likelihood by the WordPiece tokenization algorithm~\cite{Schuster2012,wu2016googles}.

\subsection{Decoding: From Tokens to Signs}
\label{subsec:From Tokens to Signs}
While the MLM task is designed to predict single tokens, in our setting, multiple signs and words may be omitted due to deterioration. 
To bridge this gap, we greedily extend the token level prediction by adapting the k-beams algorithm such that it outputs possible predictions given an Akkadian text with a sequence of missing signs. See the example at the top of Figure~\ref{fig:pipeline_figure}, where the two `x' signs in the input are predicted as \emph{a-na}. 
To achieve this, we count the number of sign delimiters (space, dot, hyphens) predicted at each time step, and choose the best $k$ candidates according to the following conditional probability:
\begin{equation}
    p(X_1,...,X_n, C)=\prod_{i=1}^np(X_i|X_1,...,X_{i-1}, C)
    \label{eq:greedy-probability}
\end{equation}
Where $X_i$ denotes the $i_{th}$ masked token, and $C$ denotes the observed context.
For example, in Figure~\ref{fig:pipeline_figure}, \emph{a-na} is composed of three sub-sign tokens: \emph{'a', '-', 'na'}, while $C = $ (\emph{`a-bat LUGAL', `a\u{s}-\u{s}ur'}), and the sequence probability is $p({na}| {-, a}, C)  \cdot p({-} | {a}, C) \cdot p({a} | C)$ .


\section{Automatic Evaluation}
\label{sec:automatic-evaluation}

We present an automatic evaluation of our models' predictions for missing signs in ancient Akkadian texts, testing several masked language modeling variants for single token prediction, as well as our greedy extension to multiple tokens and signs. In all evaluations, we mask known tokens and evaluate the model's ability to predict the original masked tokens. This setup allows us to test against large amounts of texts in Oracc from different periods of time, locations or genres.

\begin{table*}[tb!]
\resizebox{\textwidth}{!}{\begin{tabular}{c|l|cc|ccc}
\toprule
\multicolumn{1}{c}{Genre} & 
\multicolumn{1}{l}{Metric} & 
LSTM& 
\multicolumn{1}{l}{MBERT-base} & \multicolumn{1}{l}{\bertmono{}} & \multicolumn{1}{l}{MBERT+Akk} & \multicolumn{1}{l}{MBERT+Akk+Eng} \\ \midrule
\multirow{2}{*}{\begin{tabular}[c]{@{}c@{}}Royal \\ Inscription\end{tabular}}    & MRR   & .52      & .57                    & .57                         & \textbf{.83}                & \textbf{.83}                             \\
                                                                                 & Hit@5 & .60      & .65                    & .56                         & \textbf{.90}                           & \textbf{.90}                    \\ \cline{1-1}
\multirow{2}{*}{\begin{tabular}[c]{@{}c@{}}Royal or \\ Monumuental\end{tabular}} & MRR   & .51      & .61                    & .61                         & \textbf{.84}                 & .83                             \\
                                                                                 & Hit@5 & .61       & .69                    & .69                          & \textbf{.90}                           & \textbf{.90}                    \\ \cline{1-1}
\multirow{2}{*}{\begin{tabular}[c]{@{}c@{}}Astrological \\ Report\end{tabular}}  & MRR   & .53      & .55                    & .55                         & \textbf{.81}                 & .80                             \\
                                                                                 & Hit@5 & .60      & .64                    & .64                         & \textbf{.88}                          & \textbf{.88}                    \\ \midrule
\multirow{2}{*}{Lexical}                                                         & MRR   & .10      & .61                    & \textbf{.69}                & \textbf{.69}                & .66                             \\
                                                                                 & Hit@5 & .10      & .76                      & .76                         & \textbf{.85}                & \textbf{.85}                    \\ \cline{1-1}
\multirow{2}{*}{Decree}                                                          & MRR   & .49      & .67                    & .39                         & .71                         & \textbf{.74}                    \\
                                                                                 & Hit@5 & .60      & .73                      & .51                         & \textbf{.76}                & \textbf{.76}                    \\ \midrule\midrule
\multirow{2}{*}{\textbf{Overall}}                                                         & MRR   & .52      & .60                    & .50                         & \textbf{.83}                & \textbf{.83}                             \\
                                                                                 & Hit@5 & .59      & .67                    & .60                         & \textbf{.89}                & \textbf{.89}    \\\bottomrule                        
\end{tabular}}
\caption{MRR and \hit{5} precision by genre. 
The first two models from the left are our baselines:
LSTM refers to the model from \cite{fetaya2020restoration} retrained on our data,
MBERT-base refers to the \emph{zero-shot} \mbert{} model without training on Oracc.
The following three models are introduced in Section \ref{subsec:Masked language models}:
\bertmono{} is trained mono-lingually from scratch on Oracc Akkadian texts; MBERT+Akk finetunes on Oracc Akkadian texts; and MBERT+Akk+Eng is also finetuned on their English translations.  
The three genres at the top of the Table (Royal Inscription, Monumental, Astrological) are the most common in our test dataset and contain longer, more coherent texts. The two genres at the bottom (Lexical and Decree) contain tabular texts and non-contextualized, short sentences.}
\label{tab:results-table}
\end{table*}



\subsection{Models and Datasets}
We use two strong baselines:
(1) the LSTM model that was proposed by \citet{fetaya2020restoration}, and was retrained on our dataset using their default configuration;\footnote{https://github.com/DigitalPasts/Atrahasis}\textsuperscript{,}\footnote{https://github.com/DigitalPasts/Akkademia} and (2) the cased BERT-base multilingual model, without finetuning over Oracc.\footnote{https://huggingface.co/bert-base-multilingual-cased}

We compare these two baselines against 
our models, as presented in \ref{subsec:Masked language models}, trained in three configurations: 
(1) \bertmono{} refers to the reduced size BERT model, trained from scratch on the Akkadian texts from Oracc; (2) MBERT+Akk is a finetuned version of \mbert{} on the Akkadian texts, using the model's additional free tokens to encode sub-word tokens from Oracc; and 
(3) MBERT+Akk+Eng further finetunes on the English translations available in Oracc to introduce additional domain-specific signal.
We test all models against 5 different genres of Akkadian texts tagged in Oracc, masking 15\% of the tokens. The genres can be largely divided into two groups.
First, the Royal Inscription, Monumental, and Astrological Reports are the most common genres in the dataset and consist of longer coherent texts, mostly of essays and correspondence. 
Second, we test on two other genres: Lexical which consists mostly of tabular information (lists of synonyms and translations), and Decree that contains concatenated non-contextualized short sentences.

\subsection{Experimental Setup}
For all our experiments, we used a random 80\% - 20\% split for train and test (see Table~\ref{tab:amounts-table}).
For the monolingual model, we trained our reduced-parameters BERT model from scratch for 300 epochs with 4 NVIDIA Tesla M60 GPUs for 2 hours.
For the multilingual experiments, we finetuned \mbert{} for 20 epochs similarly to \citep{chau2020parsing}, with 8 NVIDIA Tesla M60 GPUs for 2-3 hours. We used the original architecture of \mbert{}, adding a masked language modeling head for prediction.
For the LSTM model of \citet{fetaya2020restoration}, we train for 200 epochs, with 1 NVIDIA Tesla M60 GPU for 68 hours.

\subsection{Metrics}
We report performance according to the \hit{k} and mean reciprocal rank (MRR) metrics, as defined below:

\begin{equation}
    MRR = \frac{1}{N}\sum_{i=1}^{N}\frac{1}{rank_i}
\end{equation}
\begin{equation}
    Hit@k = \frac{1}{N}\sum_{i=1}^{N}\mathbbm{1}_{[rank_i\leq k]}
\end{equation}

Where $N$ is the number of masked instances, $rank_i$ is the rank of the original masked token in the model's predictions, and $\mathbbm{1}$ is the indicator function.

The \hit{k} metric directly measures applicability in our target application, i.e., how likely is the correct prediction to appear if we present the user with our model's top $k$ predictions. MRR complements \hit{k} by providing a finer-grained evaluation, as the model receives partial credit in correlation with every ranking.

\begin{figure*}
\centering
  \includegraphics[width=0.9\textwidth]{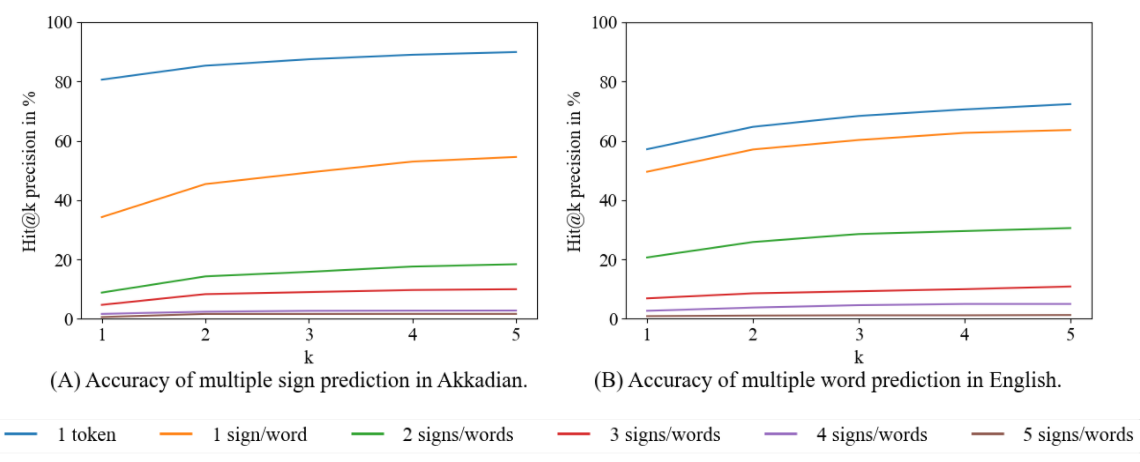}
    \caption{\label{fig:akkadian-sign-level-hit-ks}
    \hit{k} precision for sequences of varying lengths in Akkadian (A) and English (B).
    We find that both languages do well on 1 token and 1 sign, where the correct answer is expected to be in the models' top 5 predictions for half of the instances. Performance drops sharply for longer sequences, possibly due to the large search space. We directly measure the model's applicability in user studies in Section~\ref{sec:Manual-Evaluation}.
    }
\end{figure*}

\subsection{Results}
Table~\ref{tab:results-table} compares token level evaluation across our different models and genres, while Figure~\ref{fig:akkadian-sign-level-hit-ks} presents an evaluation of the prediction of multiple signs and words.
We note several interesting observations based on these results.

\paragraph{Multilingual pretraining + Akkadian finetuning achieves state-of-the-art performance.}
On average, the two \mbert{} models, which were finetuned over Oracc texts, outperform all other models by at least 20\% on both metrics.
This is particularly pronounced in the more natural first set of genres, where the multilingual models often surpass 85\% in both MRR and \hit{5}.

\paragraph{Zero-shot multilingual pretraining outperforms monolingual training.}
Surprisingly, in most tested settings, the zero-shot version of \mbert{} outperforms both \bertmono{} and the LSTM models, despite never training on Akkadian. This suggests that the signal from pretraining is stronger than that of the Akkadian texts, likely due to the relatively small amounts of data. Moreover, as \mbert{} was trained over the MLM task in other languages during its pretraining, this evaluation can be seen as a zero-shot cross-lingual transfer learning, on which \mbert{} was found to be competitive in many NLP tasks \citep{pires-etal-2019-multilingual, wu-dredze-2019-beto,conneau-etal-2020-emerging}.

\paragraph{Performance degrades on the Lexical genre.}
The gains of the multilingual models are reduced in the Lexical genre. Specifically, they are on par with \bertmono{} in this genre. This may indicate that this genre's idiosyncratic syntax does not benefit much from multilingual pretraining.

\paragraph{Context matters after finetuning \mbert{}.}
The performance of the finetuned \mbert{} is the lowest in the Decree genre and is very close to that of the MBERT-base. This is perhaps not surprising as the Decree texts are concatenations of unrelated short sentences, while one of BERT's main advantages is its learned contextualized representations of different domains.

\paragraph{Finetuning on English Oracc translations does not improve performance.}
Finetuning \mbert{} only on Akkadian (MBERT+Akk) leads to results on par with additional finetuning on English (MBERT+Akk+Eng), possibly indicating that the amount of Akkadian texts and English translations is not enough to make \mbert{} align between the two languages in Oracc's unique domains.

\paragraph{Performance degrades on longer masked sequences for both English and Akkadian.}
Figure~\ref{fig:akkadian-sign-level-hit-ks} compares our best-performing model in predicting a varying number of signs against \mbert{} on English texts, where both use our greedy decoding strategy to extend their predictions to multiple signs and words.
We note similar patterns for both languages. The performance for a single sign and word is high, and it deteriorates when more elements are predicted.
In the following section, we extend this evaluation by conducting a human evaluation that aims to test the model's applicability in a real-world setting.

\section{Human Evaluation and User Studies}

\label{sec:Manual-Evaluation}
\begin{figure}[tb!]
    \includegraphics[width=\linewidth]{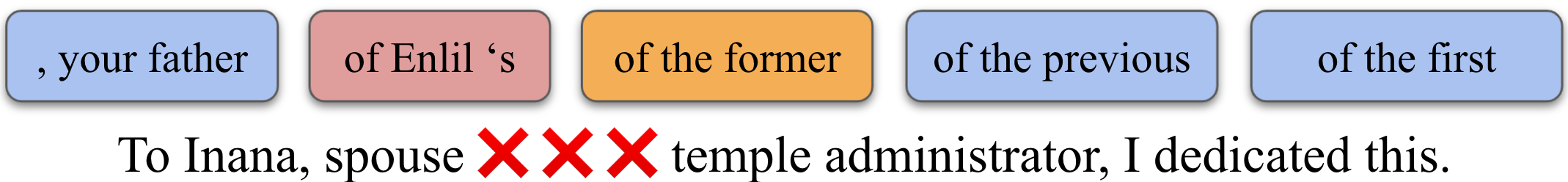}
    \noindent\rule{\columnwidth}{0pt}
    \includegraphics[width=\linewidth]{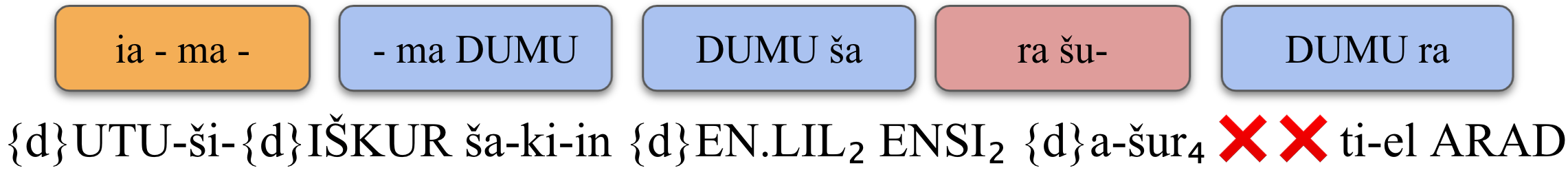}
    \caption{Human evaluation interface for English (top) and transliterated Akkadian (bottom). Given the textual context from the tablet and a missing span of text (marked by red X's), the annotator decides whether each presented option is plausible.
    The options consist of the top three model predictions (marked in blue) and two controls: the original masked span (marked in yellow) and a randomly sampled span of text functioning as a distractor (marked in red).
    }
    
    \label{fig:docanno_figure}
\end{figure}

We note that the automatic evaluation presented in the previous section offers only an upper bound of the model's ability to suggest reasonable completions, since the original text is often only one out of many other equiprobable completions of the masked text.
Consider, for example, the masked English text at the top of Figure~\ref{fig:docanno_figure}. While the original text was \example{of the former}, the model's top predictions (\example{of the previous}, \example{of the first}) may also be acceptable to scholars.
This may also explain the degradation in performance in Figure~\ref{fig:akkadian-sign-level-hit-ks}, as the number of plausible completions rises in correlation with the length of the predicted span.

To address this, we conduct a direct manual evaluation of the top performing model's predictions (\mbert{} finetuned over Oracc) in a controlled environment, on both the original Akkadian, as well as its corresponding English translation.
We begin by describing the experiment setup, which aims to cope with the inherent noise of human analysis in the MLM task, especially in an extinct language. Then, we discuss our findings, which show that the model provides sensible suggestions in most instances, while the comparison with English reveals that there is room for improvement, especially on longer sequences. 

\begin{figure}[tb!]
  \includegraphics[width=0.49\textwidth]{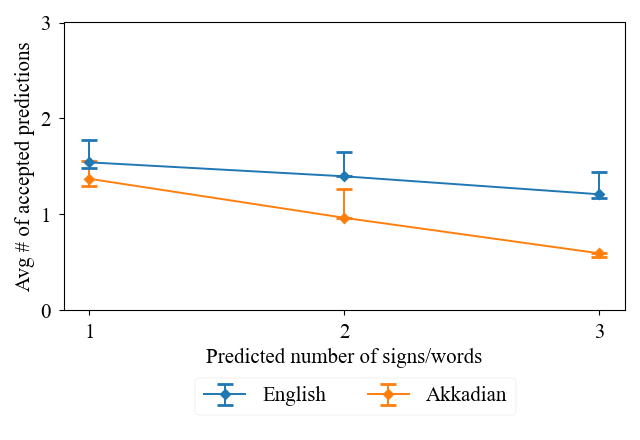}
    \caption{\label{fig:human-evaluation-akk-eng}Human evaluation results.
The X-axis represents the number of signs (in Akkadian) or words (in English) in a predicted sequence, and the Y-axis represents the average number of model predictions that our human experts approved for the given predicted sequence. The upper error bars represent false negatives, where the gold sequence was labeled not plausible. The lower error bars represent false positives, where the distractor was labeled as plausible. We find that annotators tend to introduce false negatives, while they  are less prone to falsely label distractors as plausible.}
    
\end{figure}

\subsection{Experiment Setup: Coping with Noisy Human Evaluation}
Our human evaluation of missing sign prediction in Akkadian was done by two of the authors, who are professional Assyriologists. 
They can read Akkadian at an academic level, and represent the users who work on cuneiform transliteration and may benefit from our model's predictions.
Despite their unique expertise, they do not speak the language fluently like native speakers did, and the language's natural variations over thousands of years makes the reading even more difficult.

To address this, we created an annotation scheme\footnote{Created with docanno~\cite{doccano}.} which evaluates the model's predictions and estimates the noise introduced in the annotation process.
As exemplified in Figure~\ref{fig:docanno_figure}, 
for each annotation instance, we generated 5 suggestions: 3 model predictions, the original masked term, and a distractor sequence that was randomly sampled from the Akkadian texts.\footnote{In case the model predicted the gold sequence, we added an additional model prediction, to ensure we always present 5 options.}
The annotators observe the 5 suggestions in a randomized order, oblivious to which ones are model predictions. They are then required to mark each suggestion as either plausible or implausible, given the document's surrounding context.

Inserting the original masked sequence and the distractor enabled us to quantitatively estimate two sources of noise. First, the percentage of gold samples which were marked as incorrect reflects an \emph{underestimation} of the model's ability as these are samples which in fact occurred in the original ancient texts, yet were ruled out by our experts. Similarly, the percentage of distractors marked as plausible reflects an \emph{overestimation} of the model's performance. 

By combining the estimated model accuracy~(the percentage of the predictions marked as plausible) with both sources of noise, we can estimate a range in which the actual performance of the model may lie.
Finally, for comparison with a high-resource language, we asked two fluent English speakers to annotate instances from the English translations of Oracc when predictions were generated by English BERT-base uncased model in the same experimental setup, as demonstrated at the top of Figure~\ref{fig:docanno_figure}.

We conclude this part with an example human annotation and its corresponding analysis. 

\paragraph{Annotation example.}
Consider the English annotation instance presented in Figure \ref{fig:docanno_figure}, and assume the annotator marked as plausible the following four items: the artificially introduced noise (\example{of Enlil's}); two of the model predictions:  \example{of the first}, \example{of the previous}; and the gold instance (\example{of the former}), while the remaining model prediction (\example{, your father}) is considered wrong by the human annotator.
In which case, we compute the annotator's quality assessment for this instance as $\frac{2}{3}$, while we record that they tend to \emph{overestimate} the model performance, as they marked the artificial noise as plausible. Both of these metrics~(accuracy and error estimation) are aggregated and averaged over the entire annotation.


\subsection{Results}
Each of our two annotators marked the top 5 model predictions for 70 different missing sequences, resulting in 700 binary annotations overall. 150 of these annotations were doubly annotated to compute agreement, overall finding good levels of agreement~(.81$\kappa$ for English and .79$\kappa$ for Akkadian). These were drawn from royal inscriptions, as tagged in Oracc. This genre contains straight-forward, yet elaborate syntax and is well known by our annotators.
We can make several observations based on Figure~\ref{fig:human-evaluation-akk-eng} which depicts the results of the human evaluation, based on the number of missing signs and the tested language~(Akkadian versus English). 

\paragraph{Our model's Akkadian predictions are applicably useful...}
Per sequence of one or two signs, the annotators tended to accept on average at least one suggestion as plausible, while for three signs, they accepted on average about one suggestion per two sequences. From an applicative point of view, this functionality readily lends itself to aid transliteration of missing signs for sequences of such lengths, which constitute the majority (57\%) of missing spans in Oracc.\footnote{E.g., imagine a virtual keyboard auto-complete feature that suggests plausible completions in half of the cases.} 

\paragraph{... yet performance degrades with the number of missing tokens.}
In Figure~\ref{fig:human-evaluation-akk-eng}, we observe that the performance of the Akkadian model (in orange) degrades faster than the English model (in blue) the longer the predicted sequence gets. This indicates that the greedy decoding from a single span to multiple spans works better for English than for Akkadian. Designing a better decoding scheme is left as an interesting avenue for future work.

 \paragraph{Humans tend to \emph{underestimate} the model performance.}
By examining the assessments for the artificially introduced gold and distractor sequences we can estimate that the actual model performance may be higher than our experts estimated. We see that for both languages and in most tested scenarios, our annotators were able to rule out the distractor, while they tended to also wrongly discarded the gold sequence (shown by the upper error bar), indicating that they may have also ruled out other plausible predictions made by the model.

\section{Related Work}
Most related to our work, \citet{fetaya2020restoration} designed an LSTM model which similarly aims to complete fragmentary sequences in Babylonian texts. They differ from us in two major aspects. First, they focus on small-scale highly-structured texts, for example, lists (parataxis), such as receipts or census documents~\cite{jursa2004accounting}. Second, their LSTM model does not use multilingual pretraining, instead, it is trained on monolingual Akkadian data and its parameters are randomly initialized. In Section \ref{sec:automatic-evaluation}, we retrain their model on our data, showing that it underperforms on all genres compared to models which were pretrained using multilingual data, even in a zero-shot setting, further attesting to the valuable signal of multilingual pretraining in low-resource settings.

Predating \citet{fetaya2020restoration}, \citet{assael-etal-2019-restoring} developed a model which predicts missing characters and words in ancient Greek. Similarly to \citet{fetaya2020restoration}, they train a bi-LSTM model on monolingual data. 

Other works have used Oracc and other Akkadian resources and may benefit from our language model for Akkadian. \citet{jauhiainen-etal-2019-language} used Oracc for a shared task around language and dialect identification. \citet{luukko-etal-2020-akkadian} recently introduced a syntactic treebank for Akkadian over texts from Oracc, while \citet{sahala-etal-2020-babyfst} built a morphological analyzer using annotations from Oracc. Finally, \citet{gordin2020reading} automatically transliterated Unicode cuneiform glyphs into the Latinized transliterated form.

Several recent works also noticed the cross-lingual transfer capabilities of \mbert{}.
\citet{wu-dredze-2019-beto} and \citet{conneau-etal-2020-emerging} found that \mbert{} can successfully learn various NLP tasks in a zero-shot setting using cross-lingual transfer, pointing at the shared parameters across languages as the most important factor.
\citet{pires-etal-2019-multilingual} showed that \mbert{} is capable of zero-shot transfer learning even between languages with different writing systems.

\section{Conclusions and Future Work}
We presented a state-of-the-art model for missing sign completion in Akkadian texts, using multilingual pretraining and finetuning on Akkadian texts. Interestingly, we discovered that in such a low-resource setting, the signal from pretraining may be more important than the finetuning objective. Evidently, a zero-shot model outperforms monolingual Akkadian models. Finally, we conducted a controlled user study showing the model's potential applicability in aiding human editors.

Our work sets the ground for various avenues of future work.
First, A more elaborate decoding scheme can be designed to mitigate the degradation of performance for longer masked sequences, for example by employing SpanBERT~\citep{joshi-etal-2020-spanbert} to represent the missing sequences during training and inference. 
Second, our findings suggest that an exploration of the specific utility of similar languages, e.g., Arabic or Hebrew, may yield improvements in missing sign prediction.


\section*{Acknowledgements}
We thank Ethan Fetaya and Shai Gordin for insightful discussions and suggestions and the anonymous reviewers for their helpful comments and feedback.  This work was supported in part by  a research gift from the Allen Institute for AI.

\bibliography{emnlp2020}
\bibliographystyle{acl_natbib}

\end{document}